\documentclass[conference]{IEEEtran}
\IEEEoverridecommandlockouts

\usepackage{cite}
\usepackage{amsmath,amssymb,amsfonts}
\usepackage{algorithmic}
\usepackage{multirow}
\usepackage{graphicx}
\usepackage{textcomp}
\usepackage{xcolor}
\def\BibTeX{{\rm B\kern-.05em{\sc i\kern-.025em b}\kern-.08em
    T\kern-.1667em\lower.7ex\hbox{E}\kern-.125emX}}
\begin{document}

\title{DEFormer: DCT-driven Enhancement Transformer for Low-light Image and Dark Vision\\

\thanks{* denotes equal contributions.}
\thanks{\# denotes the corresponding author.}

}

\author{
\IEEEauthorblockN{1\textsuperscript{nd} Xiangchen Yin$^*$ }
 \IEEEauthorblockA{
 \textit{Institute of Artificial Intelligence, }\\
  \textit{Hefei Comprehensive National Science Center}\\
  \textit{University of Science and Technology of China}\\
 Hefei 230601, China \\
 E-mail: yinxiangchen@mail.ustc.edu.cn}
 
  \and
  
 \IEEEauthorblockN{2\textsuperscript{nd} Zhenda Yu$^*$ }
 \IEEEauthorblockA{\textit{Anhui University}\\
  \textit{Institute of Artificial Intelligence, }\\
  \textit{Hefei Comprehensive National Science Center}\\
 Hefei 230601, China \\
 E-mail: wa22201140@stu.ahu.edu.cn}

  \and
  
 \IEEEauthorblockN{3\textsuperscript{nd} Xin Gao}
 \IEEEauthorblockA{\textit{School of Vehicle and Mobility, } \\
 \textit{Tsinghua University}\\
 Beijing 100084, China \\
 E-mail: bqt2000405024@student.cumtb.edu.cn}

   \and
  
 \IEEEauthorblockN{4\textsuperscript{nd} Xiao Sun$^{\#}$}
 \IEEEauthorblockA{ \textit{School of Computer Science and Information Engineering, } \\
 \textit{Hefei University of Technology}\\
 Hefei 230009,China\\
 E-mail: sunx@hfut.edu.cn}




}
\maketitle

\begin{abstract}
Low-light image enhancement restores the colors and details of a single image and improves high-level visual tasks. However, restoring the lost details in the dark area is still a challenge relying only on the RGB domain. In this paper, we delve into frequency as a new clue into the model and propose a DCT-driven enhancement transformer (DEFormer) framework. First, we propose a learnable frequency branch (LFB) for frequency enhancement contains DCT processing and curvature-based frequency enhancement (CFE) to represent frequency features. Additionally, we propose a cross domain fusion (CDF) to reduce the differences between the RGB domain and the frequency domain. Our DEFormer has achieved superior results on the LOL and MIT-Adobe FiveK datasets, improving the dark detection performance.
\end{abstract}


\section{Introduction}
Low-light images have many dark areas, resulting in noise, loss details and bring a negative visual experience. Low-light sence impaires the performance of high-level visual tasks decline (such as object detection\cite{lin2014microsoft, yin2023peyolo, loh2019getting}, face detection\cite{yang2020advancing, wang2022unsupervised, wang2021hla} and video task~\cite{wang2024eulermormer,wang2024frequency}). Many low-light image enhancement (LLIE) methods have recently been proposed. Based on Retinex theory\cite{land1977retinex}, several methods \cite{li2018structure, wang2013naturalness, celik2011contextual} decompose the image into reflection and illumination components and enhance them. However, traditional methods with manual feature learning have large limitations and poor generalization. Recently, several CNN-based \cite{kalwar2023gdip, zheng2023steps, wang2022local} and Transformer-based\cite{zhang2023lrt, zhang2023complementary, singh2022action} methods have been developed to address challenges in poor lighting conditions, improving dark visual tasks.

\begin{figure}[t]
\begin{center}
\begin{tabular}{ccc}
\hspace{-3.5mm}
\includegraphics[width = 0.48\linewidth]{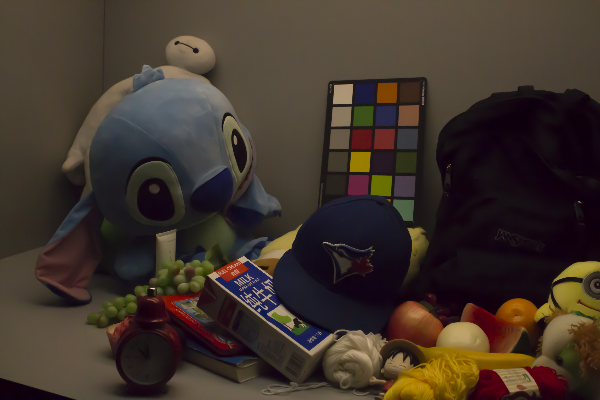} &\hspace{-5mm}
\includegraphics[width = 0.48\linewidth]{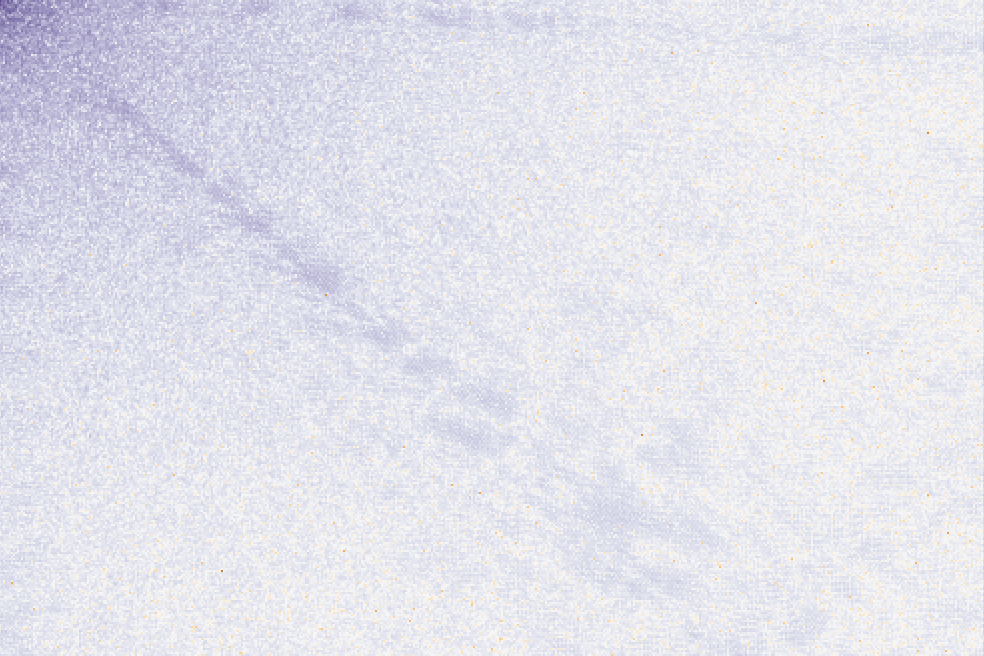} &\hspace{-5mm}
\\
(a) Input &\hspace{-4mm} (b)  Frequency spectrum 
\end{tabular}
\includegraphics[width = 0.95\linewidth]{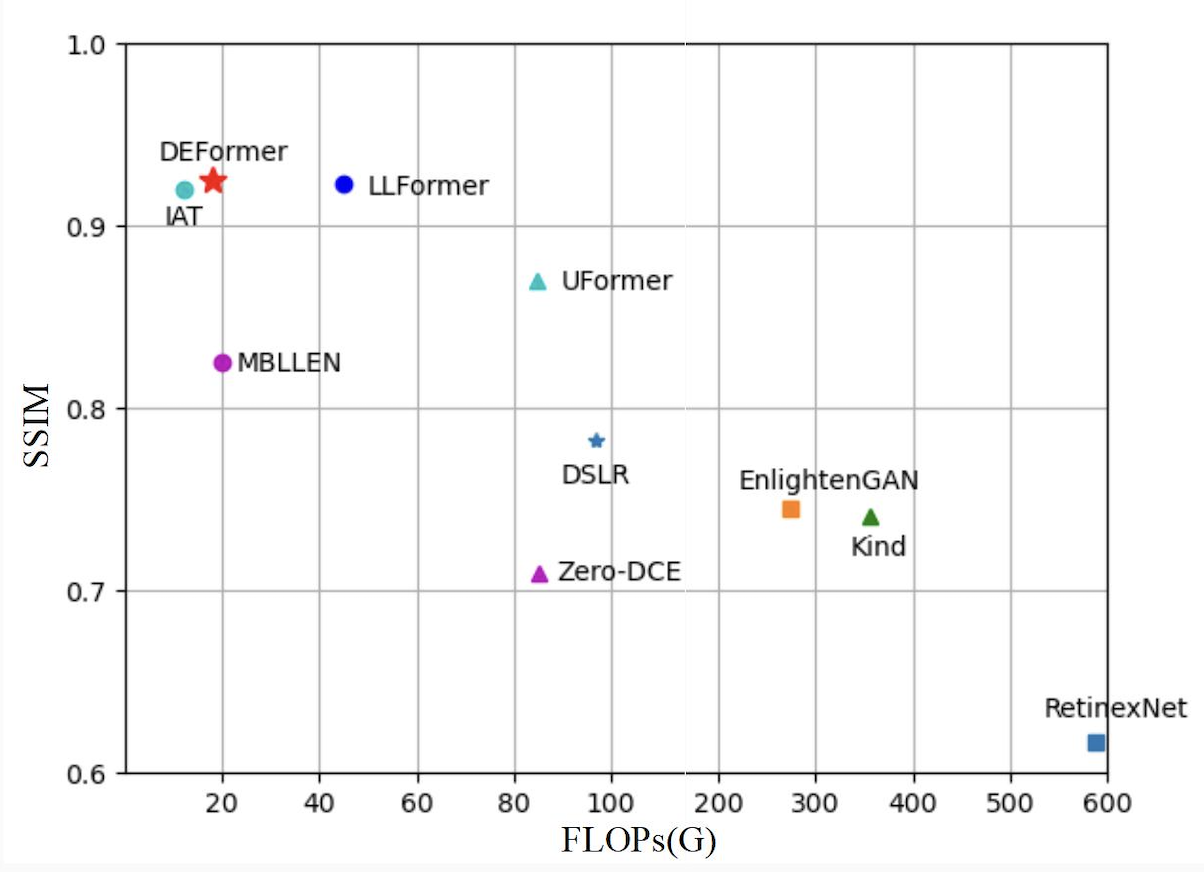}
\\
(c) Performance comparison
\end{center}

\caption{(a) represents the input, (b) represents the frequency spectrum with DCT processing. The DCT coefficient matrix concentrate the energy of the image signal. (c) represents the performance comparison between other SOTA methods on the MIT-Adobe FiveK dataset, the x-axis represents Flops and the y-axis represents SSIM. }\label{FIG1}
\end{figure}

Discrete cosine transform (DCT) is an significant part of frequency signal analysis, which has been proven effective in computer vision\cite{lin2007detection, su2024dctvit, shen2021dct}. Inspired by the frequency\cite{lee2023discrete, wang2022vtc, patro2023spectformer}, in this paper we propose a DCT-driven enhancement transformer (DEFormer) for low-light enhancement. We propose a learnable frequency branch (LFB) that contains DCT processing and curvature-based frequency enhancement (CFE) to embed information of DCT. Additionally, we propose a cross domain fusion (CDF) to reduce the differences between different domains. Extensive qualitative and quantitative experiments conducted on the LOL\cite{wei2018deep} and MIT-Adobe FiveK\cite{bychkovsky2011learning} datasets demonstrate that DEFormer achieves superior performance compared to other state-of-the-art methods. We achieve effective improvements on the ExDark\cite{loh2019getting} datasets. The performance comparison of different SOTA methods is shown in Fig. \ref{FIG1} (c). 

Our contributions are summarized as follows:
\begin{itemize}
\item Exploring the integration of DCT and Transformers, we propose a DCT-driven enhancement transformer (DEFormer) for low-light enhancement and design a learnable frequency branch (LFB) to incorporate frequency clue. 
\item We propose curvature-based frequency enhancement (CFE) to adaptively focus on frequency-band channels with rich textures. Additionally, we propose a cross domain fusion (CDF) technique to integrate features from both the RGB and frequency domains.
\item DEFormer achieved superior results on public datasets, and its end-to-end training with detectors significantly enhances performance in dark detection tasks.
\end{itemize}

\begin{figure*}
\centerline{\includegraphics[width=0.9\linewidth]{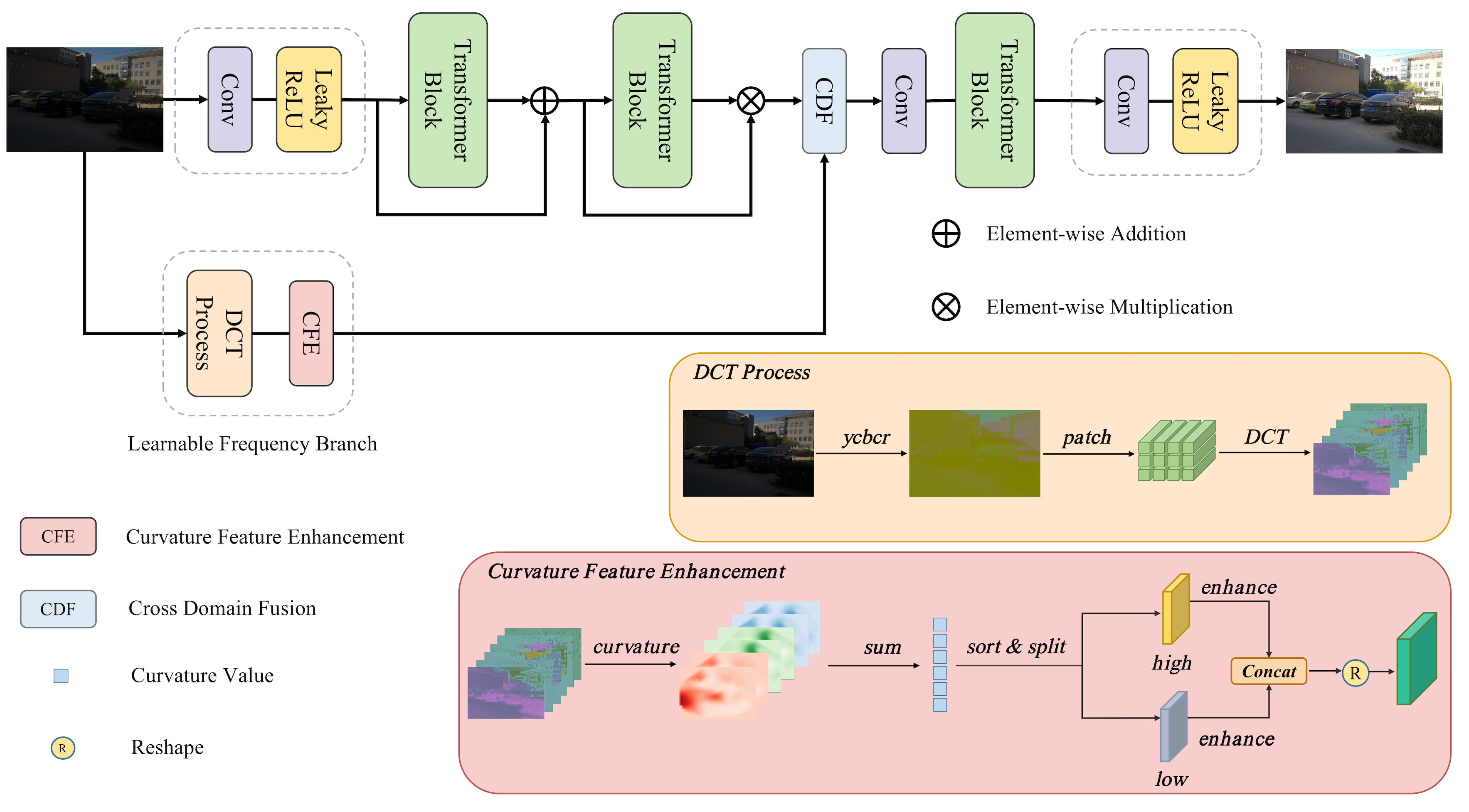}}
\caption{Overview of DEFormer. In the learnable frequency branch (LFB), we introduce frequency clues through DCT processing and curve-based frequency enhancement. In the cross domain fusion (CDF), the difference between the different domains is reduced through cross fusion. } \label{Overview}
\end{figure*}

\section{Method}
The overview of DEFormer is shown in Fig. \ref{Overview}. Giving a low-light image $I \in R^{H \times W \times 3}$ as input, we firstly extract the shallow features $F_0 \in R^{H \times W \times 16}$ through a $3 \times 3$ convolution layer. Then, we apply two transformer blocks to predict a multiple map and an add map, respectively, thereby obtaining the features of the RGB domain $F_{rgb} \in R^{H \times W \times 16}$. In the learnable frequency branch (LFB), we perform DCT processing on $I$ and divide it into high and low parts using the curvature-based frequency enhancement (CFE), enhancing each part separately, throughout tensor concat we obtain the features $F_1 \in R^{H/8 \times W/8 \times 192}$. We reshape $F_1$ to $R ^{H \times W \times 3}$ and extract features of the frequency domain $F_{f} \in R^{H \times W \times 16}$. We reduce the differences of domain between $F_{rgb}$ and $F_{f}$ in cross domain fusion (CDF). Following the fusion, we apply a transformer block to extract deep features $F_2 \in R^{H \times W \times 16}$. Finally, we use a $3 \times 3$ convolution and Leaky ReLU function to obtain the enhanced image $\hat{I} \in R^{H \times W \times 3}$.

\begin{figure}
\centerline{\includegraphics[width=1\linewidth]{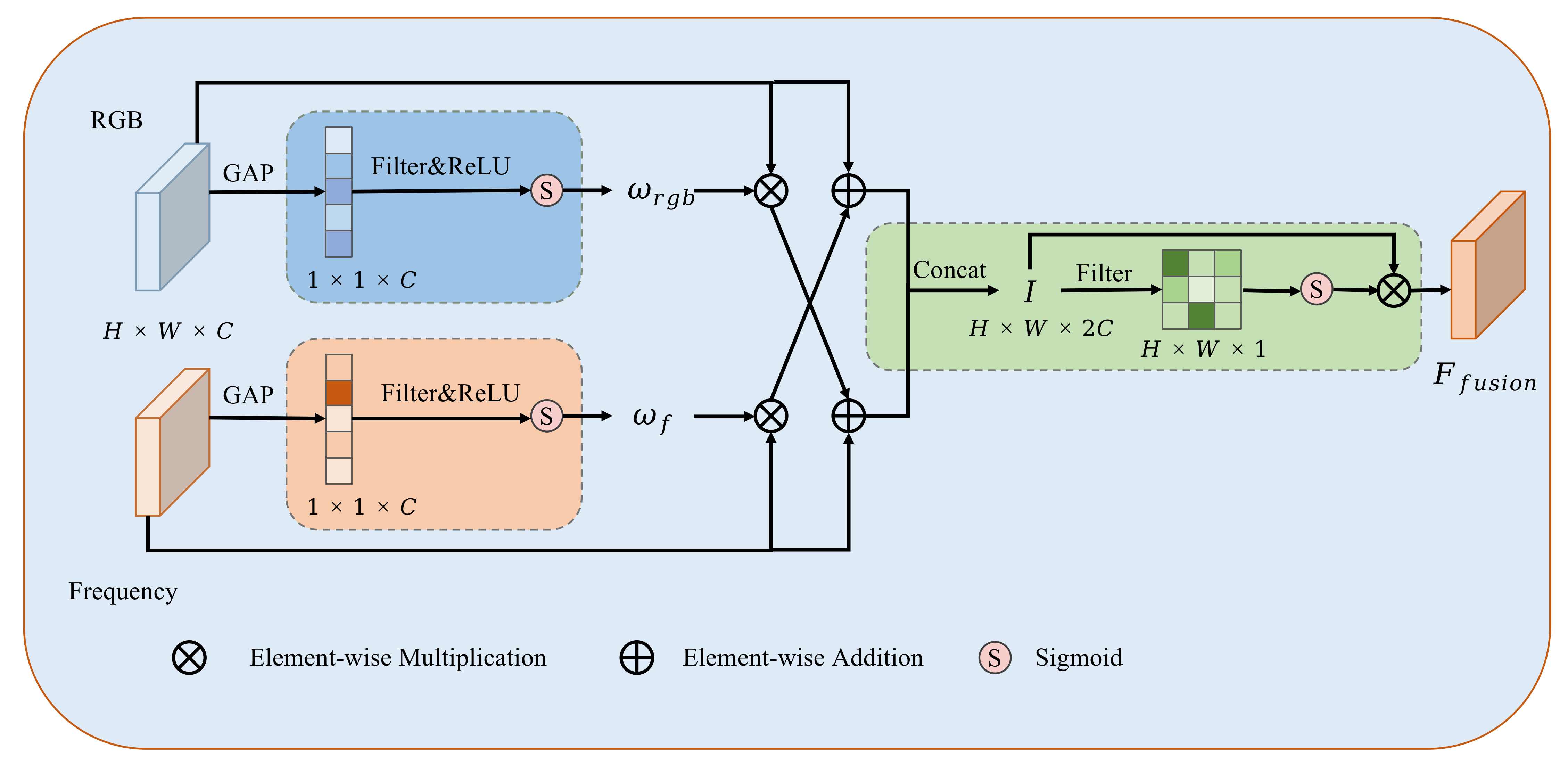}}
\caption{Details of cross domain fusion (CDF). We first complement the information between different domains through cross fusion, controling the spatial information through a soft attention to reduce noise propagation. } \label{CDF}
\end{figure}

\subsection{Learnable Frequency Branch}
Recovering lost details in dark regions is difficult with only in the RGB domain. Therefore, we apply DCT to introduce the frequency as a additional clue. The learnable frequency branch (LFB) consists of DCT processing and curvature-based frequency enhancement (CFE), as shown in Fig. \ref{Overview}.

\textbf{DCT process}. Given an RGB image $x \in R^{H \times W \times 3}$, convert it to the color space of YCbCr to get $x_{ycbcr} \in R^{H \times W \times 3}$. Then we split $x_{ycbcr}$ in each channel according to the patch of $8 \times 8$ to get $p_{i, j} \in R^{8 \times 8}, \quad 1 \leq i \leq H/8, 1 \leq j \leq W/8$. After that we process each patch through slide window DCT to obtain local frequency information. DCT has the characteristics of energy concentration, as shown in Fig. \ref{FIG1} (b), the information is concentrated in the upper left corner of the spectrum, which is also helpful for the separation of noise.
The DCT processing of the image is described as
\begin{equation}
\begin{aligned}
    F(u, v) = \frac{2}{\sqrt{HW}} \sum_{i=0}^{H-1} \sum_{j=0}^{W-1} f(i, j) \\
    cos \frac{u(2i+1) \pi}{2H}cos \frac{v(2j+1) \pi}{2W}
\end{aligned}
\end{equation}
where $1 \leq u \leq H$ and $1 \leq v \leq W$, $F(*)$ is the frequency signals and $f(*)$ is the RGB signals. We attribute all the components of the same frequency band to a channel and get the feature $x_{f} \in R^{H/8 \times W/8 \times 192}$ in the frequency domain.

\textbf{Curvature-based frequency enhancement}. In 2D surfaces, the average curvature can represent the surface roughness, which originates from the details of the image. We measure different frequency bands on the channel based on the curvature to obtain the quantization parameter of the feature on each channel. We adopt a linear method to reduce the amount of curvature calculation, this process is described as


\begin{equation}
    C \approx \omega \cdot X
\end{equation}
where $X \in R^{H \times W \times C}$ represents the frequency feature, $C \in R^{H \times W \times C}$ represents the curvature map, $\omega$ represents initial weight, adapting to the network through microscopic during the training process.
\begin{equation}
    \omega = \begin{bmatrix}
  -\frac{1}{16} & \frac{5}{16} & -\frac{1}{16} \\
  \frac{5}{16} & -1 & \frac{5}{16} \\
  -\frac{1}{16} & \frac{5}{16} & -\frac{1}{16}
    \end{bmatrix}
\end{equation}
then we sum the curvature map in two dimensions to obtain an energy vector $E \in R^{1 \times 1 \times C}$ representing each channel. We sort $E$ and divide the frequency features into high part and low part according to 3:1. In high part we apply several $\{ Conv, BN, ReLU \}$ and skip connection to enhance frequency domain information. In the low part, we use a $3 \times 3$ convolution for simple calibration and the two parts are concat in channel dimensions. Finally, we reshape the enhanced $X$ to $R^{H \times W \times 3}$ and use a convolution to get the enhanced frequency domain feature $F_{f} \in R^{H \times W \times 16}$.

\subsection{Cross Domain Fusion}
We design a cross domain fusion (CDF) to reduce the differences between the frequency domain $F_f$ and the feature of the RGB domain $F_{rgb}$, as shown in Fig. \ref{CDF}. First, we obtain global information by global average pooling along the channels and obtain attention vectors through two filters respectively. This process is described as

\begin{equation}
    \omega_{rgb} =\sigma(Filter(GAP(F_{rgb})))
\end{equation}

\begin{equation}
    \omega_{f} =\sigma(Filter(GAP(F_{f})))
\end{equation}
where $\sigma$ represents the sigmoid function, and we need to scale the weights to $[0,1]$. GAP represents global average pooling. The weights can effectively suppress the importance of the noisy part. The cross fusion is described as

\begin{equation}
    I = relu(\omega_{rgb} \cdot F_{rgb} + F_{f} || \omega_{f} \cdot F_{f} +F_{rgb})
\end{equation}
where $I$ represents the features of the initial fusion and $||$ represents the tensor concat. This process effectively reduces large differences between domains. We apply a $1 \times 1$ convolution layer and sigmoid function to obtain the information of $I$ in the spatial dimension to get a more robust fusion representation. The process is as follows

\begin{equation}
    F_{fusion} = \sigma(Filter(I)) \cdot I
\end{equation}
where $F_{fusion}$ is fused feature. 

\begin{figure}[t]
\begin{center}
\begin{tabular}{cccc}
\includegraphics[width = 0.3\linewidth]{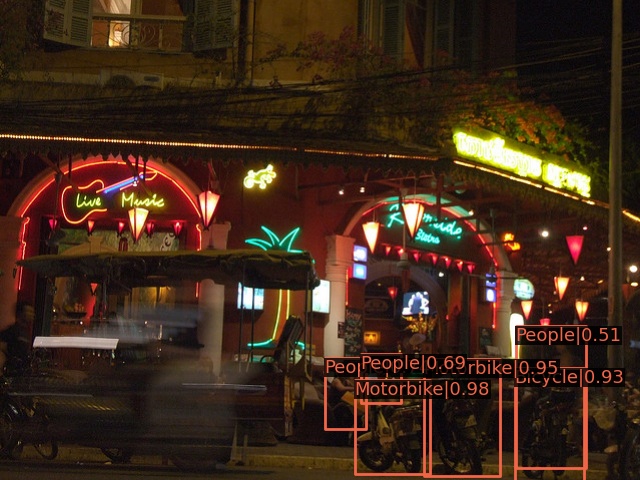} &\hspace{-3mm}
\includegraphics[width = 0.3\linewidth]{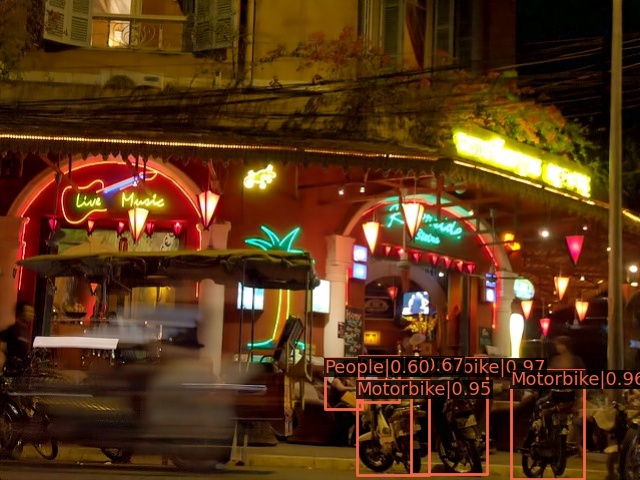} &\hspace{-3mm}
\includegraphics[width = 0.3\linewidth]{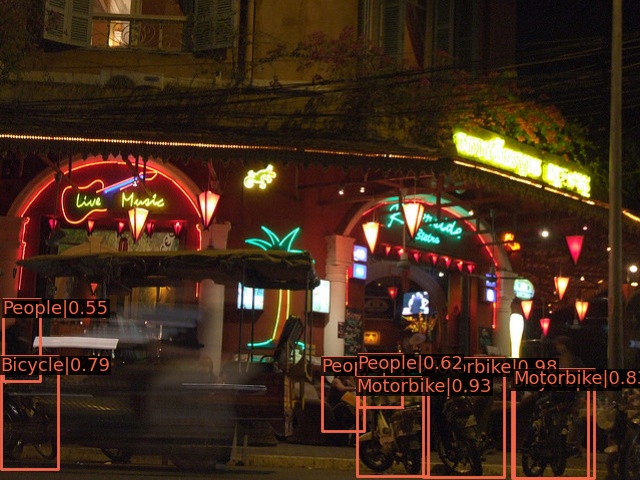} &\hspace{-3mm}
\\
YOLOv3 &\hspace{-2mm} MBLLEN &\hspace{-2mm} DEFormer
\end{tabular}
\end{center}
\caption{Visualization of dark object detection on the ExDark dataset. MBLLEN directly predict the enhanced image and apply detector training, DEFormer uses end-to-end training. }\label{FIG3}
\label{Vis_Exdark}
\end{figure}

\begin{table*}[h]
\caption{Comparison of low-light enhancement model on LOL and MIT-Adobe FiveK datasets. The first best result and the second best result is marked as blue and green respectively.}\label{TABLE 1}
\label{comparison_low}
\centering
\begin{tabular}{l|cc|cc|cc}
\hline
 \multirow{2}{*}{Model}  &  \multicolumn{2}{c|}{LOL}  &  \multicolumn{2}{c|}{MIT-Adobe FiveK} & \multirow{2}{*}{Parameters (M)} & \multirow{2}{*}{FLOPs (G)}\\ \cline{2-5}
   & PSNR & SSIM & PSNR & SSIM & & \\
\hline
RetinexNet\cite{wei2018deep}  &	16.77 &	0.562 &	 12.51 &	0.671 &	0.84 &	587.47\\
Kind\cite{zhang2019kindling}  &	20.86 &	0.790 &	14.54 &	0.741 &	8.16 &	356.72\\
Zero-DCE\cite{guo2020zero}  &	14.86 &	0.589 &	13.20 &	0.709 &	\color{blue}{0.08} &	84.99 \\
EnlightenGAN\cite{jiang2021enlightengan}  &	17.48	&0.677	& 13.26 &	0.745 &	8.64 &	273.24	\\	
IAT\cite{cui2022illumination}  & 21.87 & 0.788 & 25.10 & 0.920 & \color{green}{0.09} & \color{blue}{12.16}\\
UFormer\cite{wang2022uformer}  &	19.39 &	0.786 & 21.92	 & 0.870 & 	5.29 &	84.94\\	
Restormer \cite{zamir2022restormer}  &	22.37 &	\color{green}{0.816} &	24.92 &	0.911  & 26.13	& 281.98 \\		
LLFormer\cite{jie2023llformer}  & 	\color{green}{23.64} &	\color{green}{0.816} &	\color{blue}{25.75} &	\color{green}{0.923} &	24.55 &	45.04 \\
PairLIE\cite{fu2023learning}  & 19.51 & 0.736 & - & -	& 3.42 & 179.58\\
DEFormer (Ours) 	& \color{blue}{23.73}	& \color{blue}{0.821} & \color{green}{25.14} &	\color{blue}{0.925} &	2.25 &	\color{green}{17.96}\\

\hline
\end{tabular}
\end{table*}





\section{Experiments}
For low-light enhancement, we apply the LOL and MIT-Adobe FiveK datasets. We conduct our experiments under Ubuntu 18.04, Pytorch, RTX 3090. We use SGD and set the initial learning rate and weight decay to 0.0001 and 0.00005, respectively. During training, epoch is set to 1000, and batch-size is set to 8. We use PSNR, SSIM, parameters and Flops for evaluation. 

\begin{figure*}[t]
\begin{center}
\begin{tabular}{cccccc}
\hspace{-4mm}
\includegraphics[width = 0.16\linewidth]{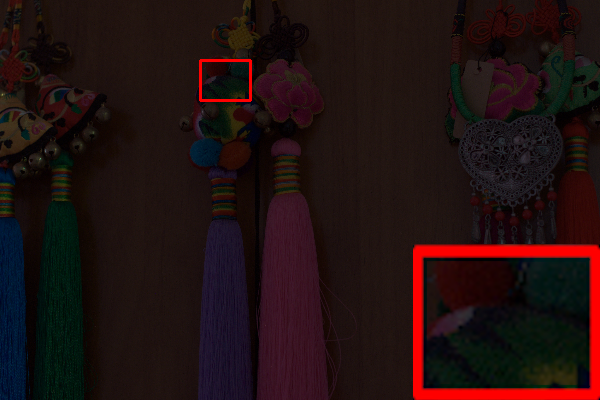} &\hspace{-5mm}
\includegraphics[width = 0.16\linewidth]{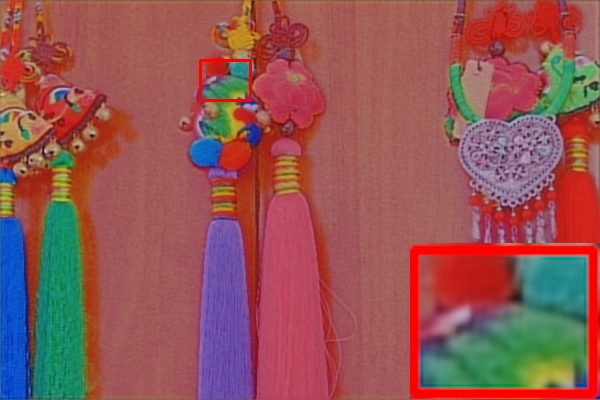}&\hspace{-5mm}
\includegraphics[width = 0.16\linewidth]{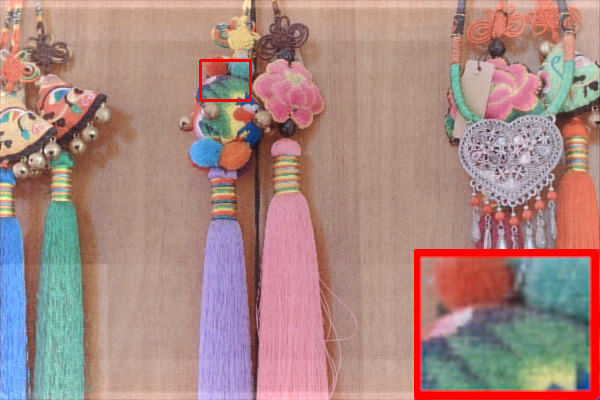}&\hspace{-5mm}
\includegraphics[width = 0.16\linewidth]{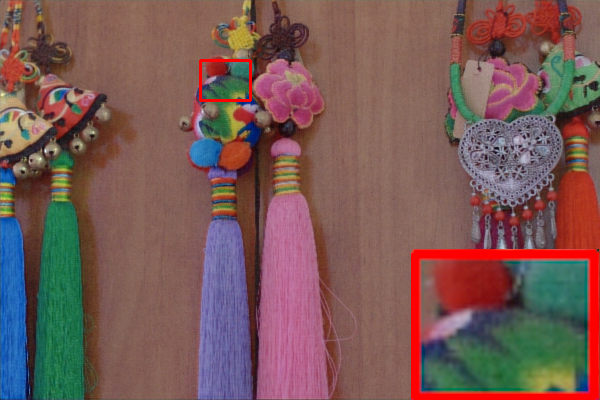}&\hspace{-5mm}
\includegraphics[width = 0.16\linewidth]{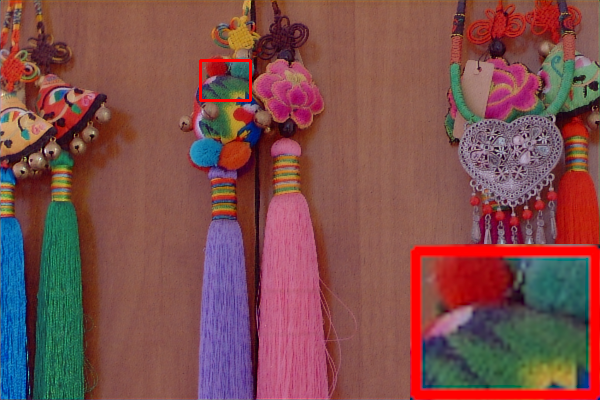}&\hspace{-5mm}
\includegraphics[width = 0.16\linewidth]{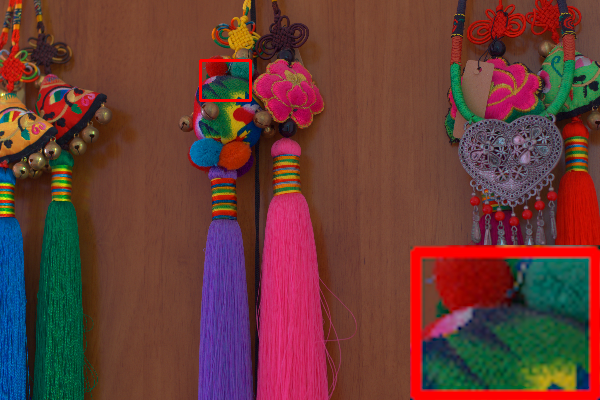}
\\
\hspace{-4mm}
\includegraphics[width = 0.16\linewidth]{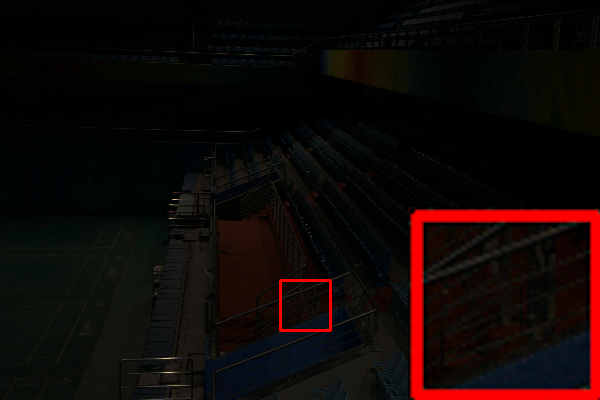}&\hspace{-5mm}
\includegraphics[width = 0.16\linewidth]{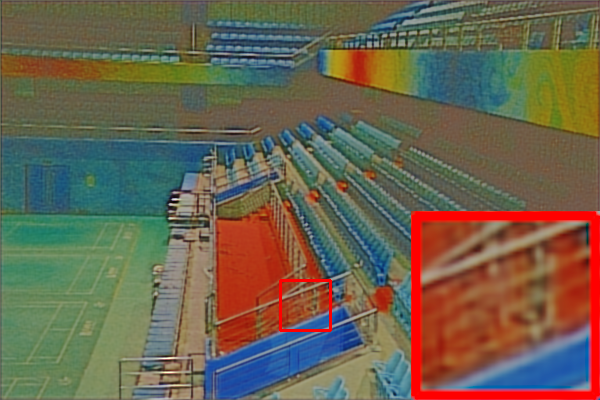}&\hspace{-5mm}
\includegraphics[width = 0.16\linewidth]{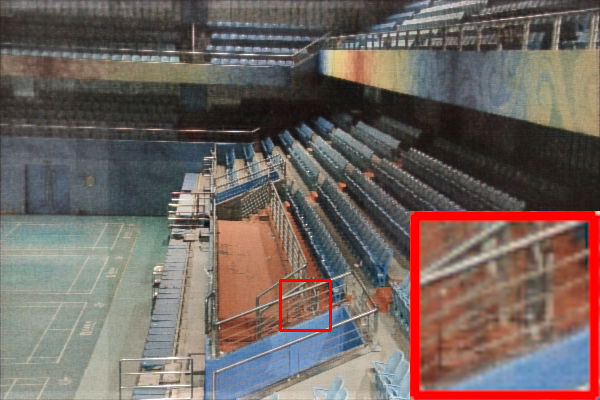}&\hspace{-5mm}
\includegraphics[width = 0.16\linewidth]{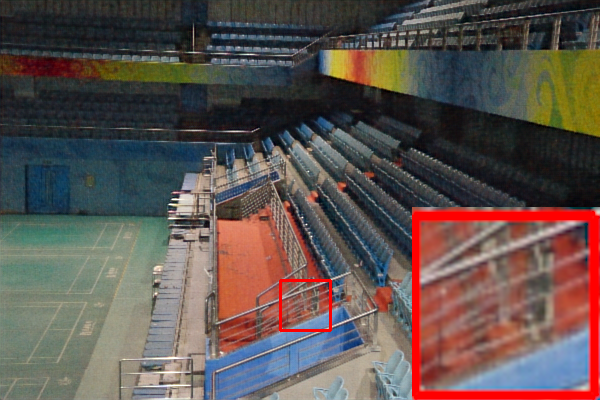}&\hspace{-5mm}
\includegraphics[width = 0.16\linewidth]{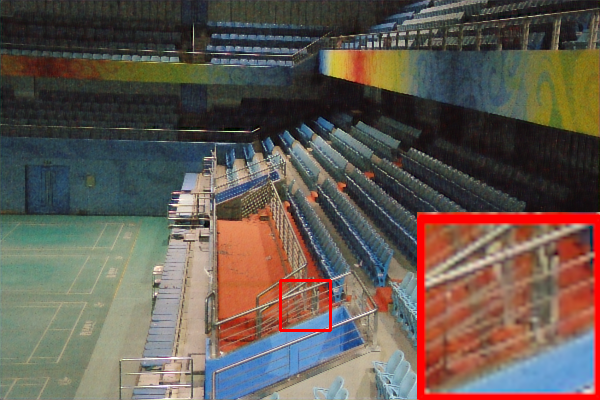}&\hspace{-5mm}
\includegraphics[width = 0.16\linewidth]{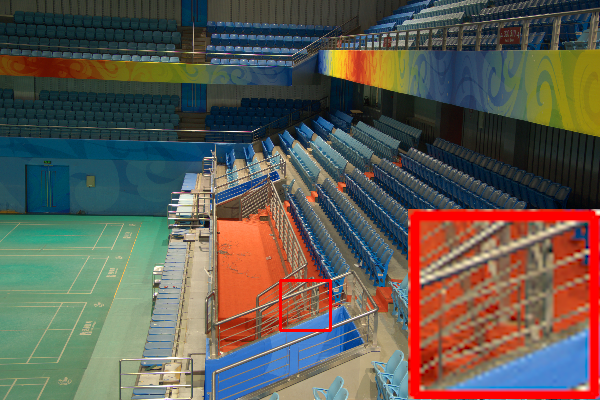}
\\
\hspace{-4mm}
\includegraphics[width = 0.16\linewidth]{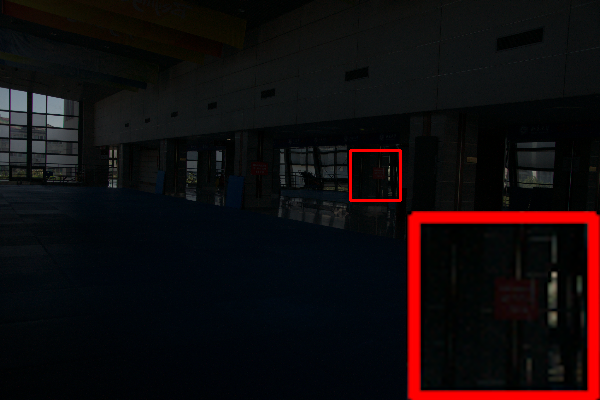}&\hspace{-5mm}
\includegraphics[width = 0.16\linewidth]{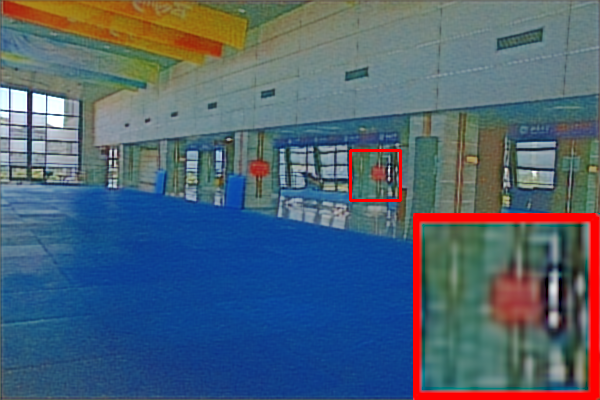}&\hspace{-5mm}
\includegraphics[width = 0.16\linewidth]{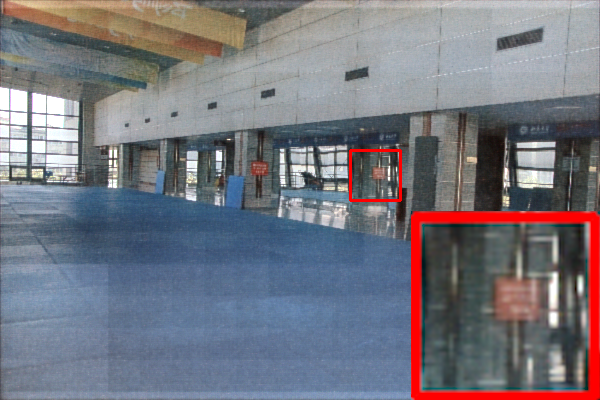}&\hspace{-5mm}
\includegraphics[width = 0.16\linewidth]{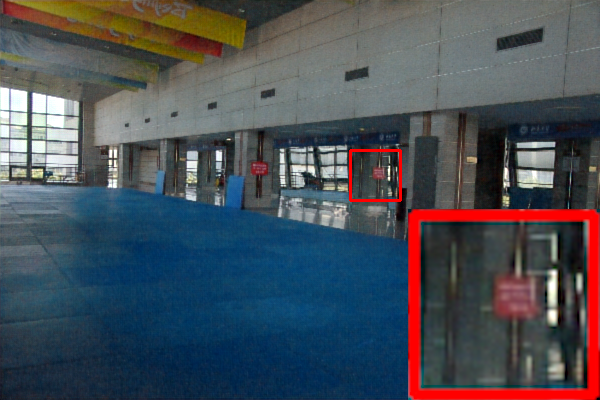}&\hspace{-5mm}
\includegraphics[width = 0.16\linewidth]{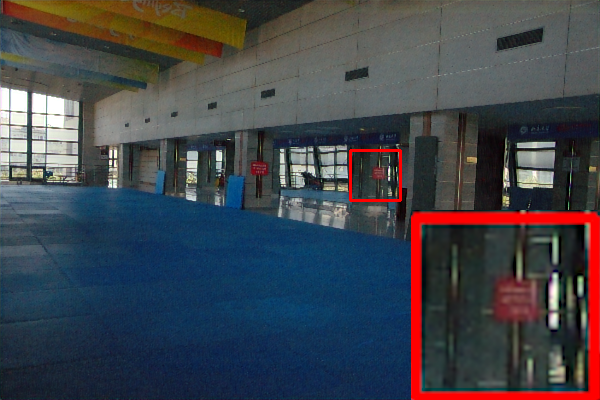}&\hspace{-5mm}
\includegraphics[width = 0.16\linewidth]{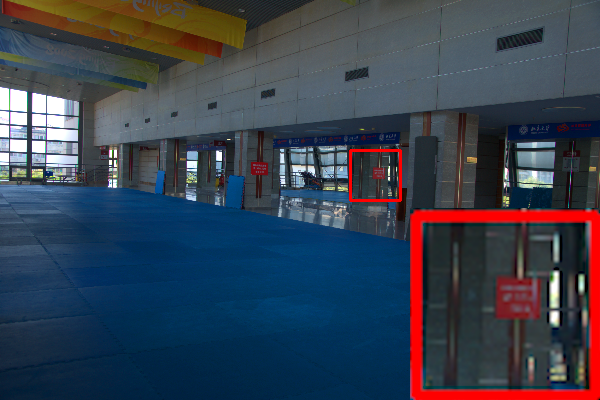}
\\
Input &\hspace{-5mm}    RetinexNet  &\hspace{-5mm}  UFormer &\hspace{-5mm} LLFormer  &\hspace{-5mm} DEFormer &\hspace{-5mm} GT
\\
\end{tabular}
\end{center}
\caption{ Visualization of different low-light image enhancement methods on the LOL dataset. Each row is a different image sample. We advise that zoom in to observe the details. }\label{FIG2}
\label{Vis-Low}
\end{figure*}

\subsection{Comparison of Low-light Enhancement}
We quantitatively compared DEFormer with other state-of-art methods such as RetinexNet \cite{wei2018deep}, Uformer \cite{wang2022uformer}, LLFormer \cite{jie2023llformer} and IAT \cite{cui2022illumination}, the results are shown in Table \ref{comparison_low}. Our DEFormer has achieved superior results on the LOL and MIT-Adobe FiveK datasets, with only 17.96G on the FLOPs. We visualized the results of several SOTA models and DEFormer on the LOL dataset, as shown in Fig. \ref{Vis-Low}, in contrast our method had high quality in both texture and color.

In addition, we fine-tuned the DEFormer and YOLOv3 detectors through end-to-end training and visualized on the ExDark datasets, as shown in Fig. \ref{Vis_Exdark}. Our DEFormer has achieved an effective improvement on the ExDark dataset, which is 2.1\% higher than the baseline in mAP. This demonstrated that our model improves downstream visual tasks.

\begin{table}[t]
\caption{Ablation study for DEFormer. "FI" represents frequency information, "CFE" represents curvature-based frequency enhancement, "CDF" represents cross domain fusion. }\label{TABLE 2}
\resizebox{\linewidth}{!}{
\label{ablation}
\centering
\begin{tabular}{l|c|cccccc}
\hline
Component  & PSNR/SSIM & FLOPs (G) & Para(M)\\
\hline
Baseline & 21.12/0.793 & 13.54 & 1.24\\
LFB(FI) & 21.57/0.797 & 13.78 &	1.28\\ 
LFB(CFE+FI) & 22.14/0.805 & 15.48 & 2.20\\
LFB+CDF & 23.73/0.821 &	17.96 &	2.25\\
\hline
\end{tabular}

}
\end{table}

\subsection{Ablation Study}
We conducted ablation studies on the LOL dataset, the results are shown in Table \ref{ablation}. We directly added the frequency domain information to the network, PSNR and SSIM increased by 0.45 and 0.004 respectively. After adopting the curvature-based frequency enhancement, PSNR and SSIM are 1.02 and 0.012 higher than the baseline, respectively. After using cross domain fusion, PSNR and SSIM have increased by 2.61 and 0.028 respectively. Overall, this ablation study demonstrated that using frequency information as additional clue in the network is an effective solution.

\section{CONCLUSIONS}
In this paper, we propose DEFormer using frequency as a new clue, which mainly includes learnable frequency branch (LFB) and cross domain fusion (CDF). Extensive experiments demonstrate that DEFormer outperforms state-of-the-art methods in low-light enhancement and improves downstream detection tasks well as preprocessing.

\section{Acknowledgments}

This work was supported by Special Project of the National Natural Science Foundation of China (62441614), Anhui Province Key R\&D Program (202304a05020068) and General Programmer of the National Natural Science Foundation of China (62376084).

\bibliographystyle{IEEEtran}
\bibliography{reference}
\end{document}